\documentclass[runningheads]{llncs}
\usepackage{graphicx}
\usepackage{comment}
\usepackage{amsmath,amssymb} 
\usepackage{color}

\usepackage{epsfig}
\usepackage{graphicx}
\usepackage{amsmath}
\usepackage{amssymb}
\usepackage{booktabs}
\usepackage{subcaption}

\usepackage[pagebackref=true,breaklinks=true,letterpaper=true,colorlinks,bookmarks=false]{hyperref}

\begin{document}
\pagestyle{headings}
\mainmatter
\def\ECCVSubNumber{3582}  

\titlerunning{Adversarial Grammar}
%
\author{
AJ Piergiovanni\inst{1} \and
 Anelia Angelova\inst{1} \and Alexander Toshev\inst{1}
\and \\Michael S. Ryoo\inst{1,2}}
\authorrunning{A. Piergiovanni et al.}
%
\institute{Robotics at Google \and
    Stony Brook University\\
\email{\{ajpiergi,anelia,toshev,mryoo\}@google.com}}

\title{Adversarial Generative Grammars for Human Activity Prediction}

\maketitle

\begin{abstract}

In this paper we propose an adversarial generative grammar model for future prediction. 
The objective is to learn a model that explicitly captures temporal dependencies,
providing a capability to forecast multiple, distinct future activities. 
Our adversarial grammar is designed so that it can learn stochastic production rules from the data distribution, jointly with its latent non-terminal representations. 
Being able to select \emph{multiple} production rules during inference leads to different predicted outcomes, thus efficiently modeling many plausible futures. 
The adversarial generative grammar is evaluated on the Charades, MultiTHUMOS, Human3.6M, and 50 Salads datasets and on two activity prediction tasks: future 3D human pose prediction and future activity prediction. The proposed adversarial grammar outperforms the state-of-the-art approaches, being able to predict much more accurately and further in the future, than prior work. 
\textbf{Code will be open sourced}.
\end{abstract}

\vspace{-8mm}
\section{Introduction}

Future prediction in videos is one of the most challenging visual tasks. Accurately predicting future activities or human pose has many important applications, e.g., in video analytics and robot action planning.
Prediction is particularly hard because it is not a deterministic process as multiple potential `futures' are possible, especially for predicting real-valued output vectors with non-unimodal distribution. Given these challenges, we address the important 
question of how the sequential dependencies in the data should be modeled and how multiple possible long-term future outcomes can be predicted at any given time.

\begin{center}
\vspace{-0.1cm}
    \includegraphics[width=0.7\linewidth]{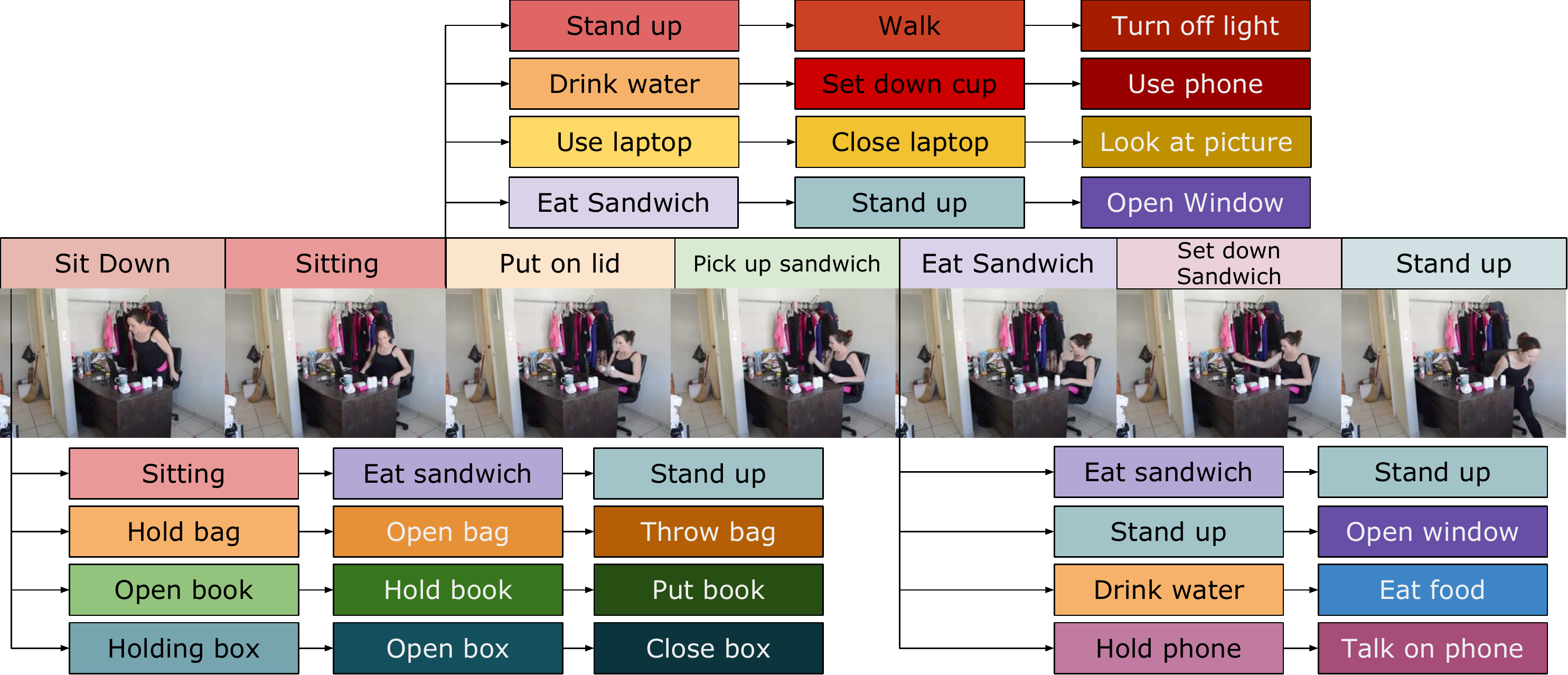}
    \captionof{figure}{The Adversarial Generative Grammar predicts future activities in videos and can generate many other plausible ones.}
    \label{fig:activity}
\vspace{0.11cm}
\end{center}



We propose an {\it Adversarial Generative Grammar} (AGG) model for future prediction. The model is a differentiable form of a regular grammar trained with adversarial sampling of various possible futures, which is able to output real-valued predictions (e.g., 3D human pose) or semantic prediction (e.g., activity classes).
Learning sequences of actions or other sequential processes with the production rules of a grammar is valuable, as it imposes temporal structural dependencies and captures relationships between latent states. 
Each (learned) production rule of a grammar model is able to take a state representation and transition to a different future state. Using multiple rules allows the model to capture multiple branching possibilities (Figure~\ref{fig:activity}). This capability makes the grammar learning unique, different from previous sequential models including many recurrent neural network (RNN) models. 

The main technical contribution of this work is the introduction of adversarial learning approach for differentiable grammar models. This is essential, as the adversarial process allows the grammar model to produce multiple candidate future sequences that follow a similar distribution to sequences seen in the data. 
A brute force implementation of differentiable grammar learning would need to enumerate all possible rules and generate multiple sequence branches (exponential growth in time) to consider multiple futures. Our adversarial stochastic sampling process allows for much more memory- and computationally-efficient learning without such enumeration. Additionally, unlike other techniques for future generation (e.g., autoregressive RNNs), we show the adversarial grammar is able to learn longer sequences, can handle multi-label settings, and predict much further into the future.


To our knowledge, AGG is the first approach of adversarial grammar learning.
It enables qualitatively and quantitatively better solutions - ones able to successfully produce multiple feasible long-term future predictions for real-valued outputs.
The proposed approach is driven entirely by the structure imposed from learning grammar rules and adversarial losses -- i.e., no direct supervised loss is used for the grammar model training. 

The proposed approach is evaluated on different future activity prediction tasks: (i) on future action prediction -- multi-class classification and multi-class multi-label problems and (ii) on 3D human pose prediction, which predicts the 3D joint positions of the human body in the future. The proposed method is tested on four challenging datasets: Charades, MultiTHUMOS, 50 Salads, and Human3.6M.
It outperforms previous state-of-the-art methods, including RNNs, LSTMs, GRUs, grammar and memory based methods. 






\section{Related work}

\textbf{Grammar models for visual data.} The notion of grammars in computational science was introduced by \cite{chomsky1956three} for description of language, and has found a widespread use in natural language understanding.
In the domain of visual data, grammars are used to parse images of scenes~\cite{zhu2007stochastic,zhao2011image,han2008bottom}. In their position paper, \cite{zhu2007stochastic} present a comprehensive grammar-based language to describe images, and propose MCMC-based inference. 
More recently, a recursive neural net based approach was applied to parse scenes by \cite{socher2011parsing}. However, these previous works either use a traditional symbolic grammar formulation or use a neural network without explicit representation of grammar.
In the context of temporal visual data, grammars have been applied to activity recognition and parsing~\cite{moore2002recognizing,ryoo2006recognition,vo2014stochastic,pirsiavash2014parsing} but not to prediction or generation. \cite{qi2017predicting} used traditional stochastic grammar to predict activities, but only within 3 seconds. 

\textbf{Generative models for sequences}. Generative Adversarial Networks (GANs) are a very powerful mechanism for  data generation by an underlying learning of the data distribution through adversarial sampling~\cite{goodfellow2014gans}. GANs have been very popular for image generation tasks~\cite{denton2015deep,isola2017i2i,wang2018pix2pixHD,brock2019large}.
Prior work on using GANs for improved sequences generation ~\cite{yu2017seqgan,fedus2018maskgan,hu2017toward} has also been successful. Fraccaro et al. \cite{fraccaro2016sequential} proposed a stochastic RNN which enables generation of different sequences from a given state. However, to our knowledge, no prior work explored end-to-end adversarial training of formal grammar as we do. Qi et al. \cite{qi2018parser} showed a grammar could be used for future prediction, and our work builds on this by learning the grammar structure differntiably from data. 

\textbf{Differentiable Rule Learning} Previous approaches that address 
differentiable rule or grammar learning are most aligned to our work~\cite{yang2017differentiable}. 
Unlike the prior work, we are able to handle larger branching factors and demonstrate successful results in real-valued output spaces, benefiting from the adversarial learning.

\textbf{Future pose prediction.} Previous approaches for human pose prediction \cite{fragkiadaki2015recurrent,ionescu2014Human3.6,tang2018longterm} are relatively scarce. The dominant theme is the use of recurrent models (RNNs or GRUs/LSTMs)~\cite{fragkiadaki2015recurrent,martinez2017on}. Tang et al. \cite{tang2018longterm} use attention models specifically to target long-term predictions, up to 1 second in the future. Jain et al. \cite{jain2016structural} propose a structural RNN which learns the spatio-temporal relationship of pose joints. The above models, contrary to ours,  
cannot produce multiple futures, making them limited for long-term anticipation.
These results are only within short-term horizons and the produced sequences often `interpolate' actual data examples. 
Although our approach is more generic and is not limited to just pose forecasting, we show that it is able to perform successfully too on this task, outperforming others.



\textbf{Video Prediction.} Our approach is also related to the video prediction literature~\cite{finn2016unsupervised,denton2018stochastic,babaeizadeh2017stochastic,lee2018stochastic}, but more in-depth survey is beyond the scope of this work.


\begin{figure*}
  \centering
   \includegraphics[width=0.87\linewidth]{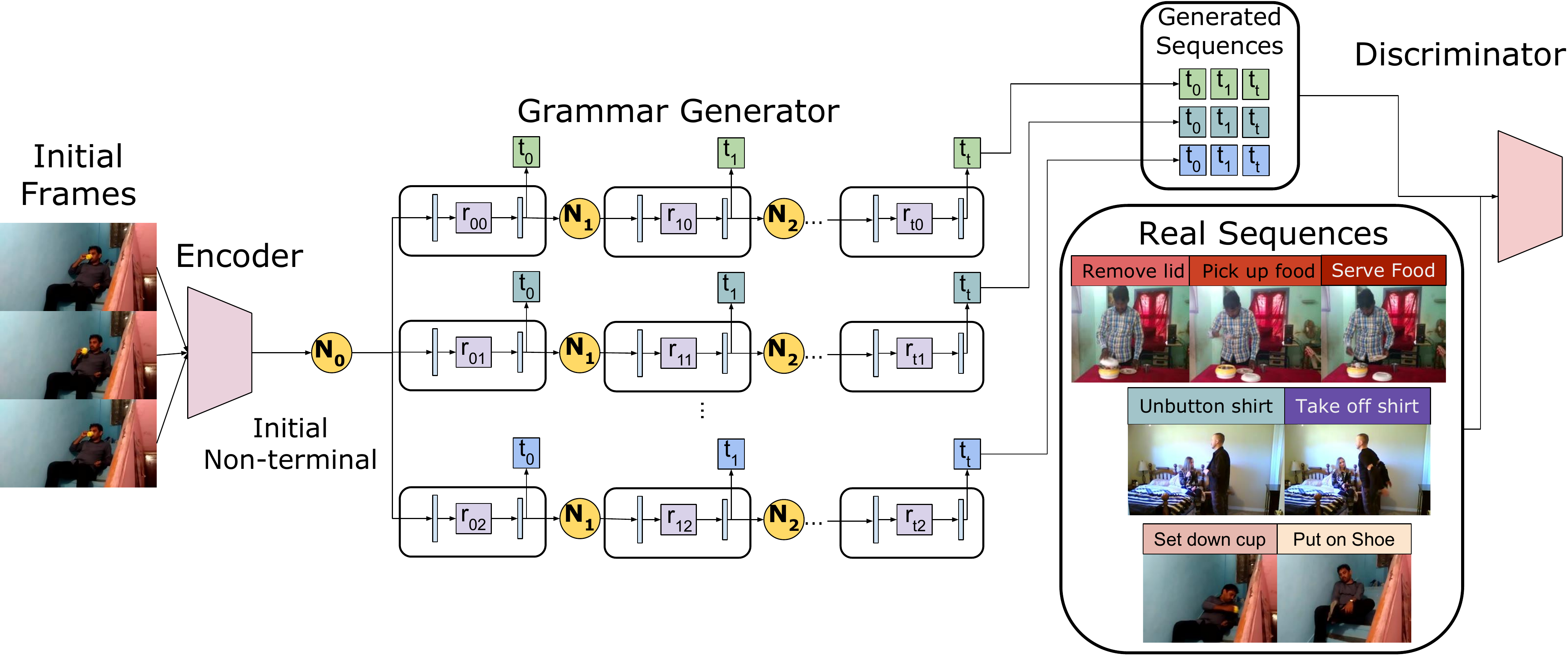}%
  \caption{Overview of the adversarial grammar model. The initial non-terminal is produced by an encoder based on the input video. The grammar then generates multiple possible sequences from the non-terminal. The generated and real sequences are used to train the adversarial discriminator, evaluating whether the generated sequences match the distribution of real sequences. 
  }
  \label{fig:overview}
\end{figure*}


\section{Approach}
\label{sec:approach}


We first introduce a differentiable form of a formal grammar, where its production rules are implemented with fully-differentiable functions to be applied to non-terminals and terminals represented with latent vectors (Section \ref{subsec:grammar}). Unlike traditional grammar induction with symbolic representations, our approach allows joint learning of latent representations and differentiable functions with the standard back-propagation. Next, we present the adversarial grammar learning approach that actually enables training of such functions and representations without spending an exponential amount of memory and computation (Sec. \ref{subsec:adversarial}).
Our adversarial grammar is trained to generate multiple candidate future sequences.
This enables robust future prediction, which, more importantly, can easily generate multiple realistic futures.

We note that the proposed approach, based on stochastic sequence learning, is driven entirely by the adversarial losses which help model the data distribution over long sequences. 
That is, while direct supervised losses can be used, we implement our approach with adversarial losses only, which learn the underlying distribution. All experiments below demonstrate the success of this approach, despite being more challenging.




\subsection{Preliminaries}
\label{subsec:prelim}

A formal regular grammar is represented as the tuple $(\mathcal{N}, \mathcal{T}, \mathcal{P}, N_0)$ where $\mathcal{N}$ is a finite non-empty set of non-terminals, $\mathcal{T}$ is a finite set of terminals (or output symbols, e.g., here actions), $\mathcal{P}$ is a set of production rules, and $N_0$ is the starting non-terminal symbol, $N_0 \in \mathcal{N}$. Production rules in a regular grammar are of the form $A\rightarrow aB$, $A\rightarrow b$, and $A\rightarrow \epsilon$, where $A,B \in \mathcal{N}$, $a, b \in \mathcal{T}$, and $\epsilon$ is the empty string. Autoregressivly applying production rules to the non-terminal generates a sequence of terminals. Note that we only implement rules of form $A\rightarrow aB$ in our grammar, allowing it to generate sequences infinitely and we represented $N$ as a real-valued vector.


Our objective is to learn such non-terminals $\mathcal{N}$ and terminals $\mathcal{T}$ as latent vector representations directly from training data, and model the production rules $\mathcal{P}$ as a (differentiable) generative neural network function. That is, the goal is to learn a nonlinear function $G$ that maps a non-terminal to a \emph{set} of (non-terminal, terminal) pairs; here $G$ is a neural network with learnable parameters. 
\begin{equation}
    G: \mathcal{N}\rightarrow \{(\mathcal{N},\mathcal{T})\}
\end{equation}
Note that this is a mapping from a single non-terminal to multiple (non-terminal, terminal) pairs. The selection of different rules enables modeling of multiple different sequences, generating different future outcomes, unlike existing deterministic models (e.g., RNNs).

The learned production rules allow modeling of the transitions between continuous events in time, for example 3D human pose or activities, 
which can naturally spawn into many possible futures at different points similarly to switching between rules in a grammar. 
For example, an activity corresponding to `walking' can turn into `running' or `stopping' or continuing the `walking' behaviour.

More formally, for any latent non-terminal $N \in \mathcal{N}$, the grammar production rules are generated by applying the function $G$ (a sequence of fully connected layers), to $N$ as:
\begin{equation} \label{eq:production}
G(N) = \{(N_i, t_i)\}_{i=1:K},
\end{equation} 


\noindent where each pair corresponds to a particular production rule for this non-terminal:
\begin{equation}
\begin{aligned}
&N\rightarrow t_1 N_1   \\
&N\rightarrow t_2 N_2  \ldots \\
&N\rightarrow t_K N_K, \text{ where } N_1, N_2,\ldots N_K \in \mathcal{N}, t_1, t_2,\ldots t_K \in \mathcal{T}, \text{ for } K \text{ rules}.
\end{aligned}
\end{equation}
This function is applied recursively to obtain a number of output sequences, similar to prior recurrent methods (e.g., RNNs such as LSTMs and GRUs). However, in RNNs, the learned state is required to abstract multiple potential possibilities into a single representation, as the mapping from the state representation to the next representation is deterministic. As a result, when learning from sequential data with multiple possibilities, standard RNNs tend to learn states as a mixture of multiple sequences instead of learning more discriminative states. By learning explicit production rules, our states lead to more salient and distinct predictions which can be exploited for learning long-term, complex output tasks with multiple possibilities, as shown later in the paper.



\subsection{Learning the starting non-terminal}

Given an initial input data sequence (e.g., a short video or pose sequences), we learn to generate its corresponding starting non-terminal $N_0$ (i.e., root node). This is used as input to $G$ so as to generate a sequence of terminal symbols starting from the given non-terminal.
Concretely, given an input sequence $X$, a function $s$ (a CNN) is learned which gives the predicted starting non-terminal:
\begin{equation}
    N_0=s(X).
\end{equation}

Notice that the function $s(X)$ serves as a jointly-learned blackbox parser that is able to estimate the non-terminal corresponding to the current state of the model, allowing future sequence generation to start from such non-terminal.


\subsection{Grammar learning}
\label{subsec:grammar}

Given a starting non-terminal, the function $G$ is applied recursively 
to obtain the possible sequences where $j$ is an index in the sequence and $i$ is one of the possible rules: 
\begin{equation}
\begin{cases}
  G(N_0) = \{(N^{1}_i, t^{1}_i)\}_i, &\text{   } j = 0  \\[8pt]
  G(N^{j}) = \{(N^{j+1}_i, t^{j+1}_i)\}_i , &\text{ for  } j > 0 
\end{cases}
\end{equation}




For example, suppose $W$ is the non-terminal that encodes the activity for `walking' sequences. Let $walking$ denote the terminal of a grammar. An output of the rule $W\rightarrow  walking W$ will be able to generate a sequence of continual `walking' behavior. 
Additional rules, e.g., $W\rightarrow  stopping U$, $W\rightarrow running V$, can be learned, allowing for the activity to switch to `stopping' or `running' (with the non-terminals $U, V$ respectively learning to generate their corresponding potential futures, e.g. `sitting down', or `falling'). Clearly, for real valued outputs, such as 3D human pose, the number and dimensionality of the non-terminals required will be larger. We also note that the non-terminals act as a form of memory, capturing the current state with the Markov property.

To accomplish the above task, $G$ (in Eq. \ref{eq:production}) has a special structure. 
$G$ takes an input of $N\in \mathcal{N}$, then using several nonlinear transformations (e.g., fully connected layers with activation functions), maps $N$ to a binary vector $r$ corresponding to a set of rules: $r = f_R(N)$. Here, $r$ is a vector with the size $|\mathcal{P}|$ whose elements specify the probability of each rule given input non-terminal. We learn $|\mathcal{P}|$ rules which are shared globally, but only a (learned) subset are selected for each non-terminal as the other rule probabilities are zero. 
This is conceptually similar to using memory with recurrent neural network methods~\cite{yagatama2018memory}, but the main difference is that the rule vectors are used to build grammar-like rule structures which are more advantageous in explicitly modeling of temporal dependencies. 


In order to generate multiple outputs, the candidate rules, $r$ are followed by the Gumbel-Softmax function~ \cite{jang2016categorical,maddison2016concrete}, which allows for stochastic selection of a rule. This function is differentiable and samples a single rule from the candidate rules based on the learned rule probabilities. The probabilities are learned to model the likelihood of each generated sequence, and this formulation allows the `branching' of sequence predictions as the outcome of the Gumbel-Softmax function differs every time, following the probability distribution.

For each given rule $r$, two nonlinear functions $f_T(r)$ and $f_N(r)$ are then learned, so that they output the resulting terminal and non-terminal for the rule $r$: $N_{new} = f_N(r)$, $t_{new} = f_T(r)$. These functions are both implemented as a sequence of fully-connected layers followed by a non-linear activation function (e.g., softmax or sigmoid depending on the task). 
The schematic of $G$ is visualized in Figure~\ref{fig:overview}, and more details on the functions are provided in the later sections.


The non-terminals and terminals are modeled as sets of high dimensional vectors with pre-specified size and are learned jointly with the rules (all are tunable parameters and naturally more complex datasets require larger capacity). For example, for a $C$-class classification problem, the terminals are represented as $C$-dimensional vectors matching the one-hot encoding for each class.



\paragraph{Difference to stochastic RNNs.} Standard recurrent models have a deterministic state given some input, while the grammar is able to generate multiple potential next non-terminals (i.e., states). This is particularly important for multi-modal state distributions. Stochastic RNNs (e.g., \cite{fraccaro2016sequential}) address this by allowing the next state to be stochastically generated, but this is difficult to control, as the next state now depends on a random value. In the grammar model, the next non-terminal is sampled randomly, but from a set of fixed candidates while following the learned probability distribution. By maintaining a set of  candidates, the next state can be selected randomly or by some other method (e.g., greedily taking most probable, beam search, etc.), giving more control over the generated sequences.

\subsection{Adversarial grammar learning}
\label{subsec:adversarial}

The function $G$ generates a set of (non-terminal, terminal) pairs, which is applied recursively to the non-terminals, resulting in new production rules and the next sets of (non-terminal, terminal) pairs.
Note that in most cases, each rule generates a different non-terminal, thus sampling $G$ many times will lead to a variety of generated sequences.
As a result, an exponential number of sequences will need to be generated during training, to cover the possible sequences, and enumerating all possible sequences is computationally prohibitive beyond $k=2$.\footnote{For a branching factor of $k$ rules per non-terminal with a sequence of length $L$, there are in $k^L$ terminals and non-terminals (for $k=2$, $L=10$ we have $\sim$1000 and for $k=3$ $\sim$60,000.} This restricts the tasks that can be addressed to ones with lower dimensional outputs because of  memory limits. When $k=1$, i.e. when there is no branching, we have an RNN-like model, unable to generate multiple possible future sequences (we also tested this in ablation experiments below). 


\noindent \textbf{Stochastic Adversarial Sampling.}
We address this problem by using stochastic adversarial rule sampling.
Given the non-terminals, which effectively contain a number of potential `futures', we use \textit{an adversarial-based sampling}, similar to GAN approaches~\cite{goodfellow2014gans}, which learns to sample the most likely rules for the given input (Figure~\ref{fig:overview}). The use of a discriminator network allows the model to generate realistic sequences that may not exactly match the ground truth (but are still realistic) without being penalized.

\textbf{Generator:}
We use the function $G$, which is the function modeling the learned grammar described above, as the {\it  generator function}.

\textbf{Discriminator:}
We build an additional {\it discriminator function} $D$.  Following standard GAN training, the discriminator function returns a binary prediction which discriminates examples from the data distribution vs. generated ones. 
Note that the adversarial process is designed to ultimately generate terminals, i.e., the final output sequence for the model. $D$ is defined as:
\begin{equation}
 p = D(t, n)     
\end{equation}
\noindent where $t = t_0t_1t_2 \ldots t_L$ is the input sequence of terminals, $n=N_0N_1N_2 \ldots N_L$ is the sequence of non-terminals ($L$ is the length of the sequence) and
$p \in [0,1]$ and reflects when the input sequence of terminals is from the data distribution or not. 
Note that our discriminator is also conditioned on the non-terminal sequence ($n=N_0N_1N_2 \ldots N_L$), thus the distribution of non-terminals is learned implicitly as well.

The discriminator function $D$ is implemented as follows: given an input sequence of non-terminals and terminals, we apply several 1D convolutional layers to the terminals and non-terminals, then concatenate their representations followed by a fully-connected layer to produce the binary prediction (see the supp. material).



\textbf{Adversarial Generative Grammar (AGG).}
The discriminator and generator (grammar) functions are trained to work jointly, generating sequences which match the data distribution. 
The optimization objective is defined as: 
\begin{equation}
\begin{aligned}
    \min_G \max_D ~&E_{x\sim p_{data}(x)} [\log D(x)] ~+ \\
    &E_{z\sim s(X)} [\log(1-D(G(z)))]
\end{aligned}
\label{eq:det_losses}
\end{equation} 
where $p_{data}(x)$ is the real data distribution and $G(z)$ is the generated sequence from an initial state based on a sequence of frames ($X$).
That is, the fist part of the loss works on sequences of actions or human pose, whereas the second works over generated sequences ($s(X)$ is the video embedding, or starting non-terminal).


Alternatively, the sequences generated by $G$ could be compared to the ground truth to compute a loss during training (e.g., maximum likelihood estimation), however, doing so requires enumerating many possibilities in order learn multiple, distinct possible sequences. Without such enumeration, the model converges to a mixture representing possible sequences from the data distribution. By using the adversarial training of $G$, our model is able to generate sequences that match the distribution observed in the dataset. This allows for computationally feasible learning of longer, higher-dimensional sequences.









\noindent\textbf{Architecture details.} The functions $G$, $f_N$ and $f_t$, $f_R$ are implemented as networks using several fully-connected layers. The detailed architectures depend on the task and dataset, and we provide them in the supplemental material.
For the pose forecasting, the function $s$ is implemented as a two-layer GRU module~\cite{cho2014GRU} followed by a 1x1 convolutional layer with $D_N$ outputs to produce the starting non-terminal. For activity prediction, $s$ is implemented as two sequential temporal 1D convolutional layers which produce the starting non-terminal.  

\section{Experiments}



We conduct experiments on two sets of problems for future prediction: future 3D human pose forecasting and future activity prediction. The experiments are done on four public datasets and demonstrate strong performance of the proposed approach over the state-of-the-art and the ability to produce multiple future outcomes, to handle multi-label datasets, and to predict further in the future than prior work.

\begin{table*}  [h!]
\small
\vspace{-0.9cm}
  \centering
       \caption{Evaluation of future pose for specific activity classes. Results are Mean Angle Error (lower is better). Human3.6M dataset.}
  {

  \begin{tabular}{l|c|c|c|c|c|c|c|c}
  \hline

Methods &Walking & & & & & & &  \\
\hline
    &80ms &160ms &320ms &400ms &560ms &640ms &720ms &1000ms\\
ERD \cite{fragkiadaki2015recurrent} &0.77 &0.90 &1.12 &1.25 &1.44 &1.45 &1.46 &1.44  \\
LSTM-3LR \cite{fragkiadaki2015recurrent} &0.73 &0.81 &1.05 &1.18 &1.34 &1.36 &1.37 &1.36 \\
Res-GRU \cite{martinez2017on} &0.27 &0.47 &0.68 &0.76 &0.90 &0.94 &0.99 &1.06 \\
Zero-velocity \cite{martinez2017on} &0.39 &0.68 &0.99 &1.15 &1.35 &1.37 &1.37 &1.32 \\
MHU \cite{tang2018longterm} &0.32 &0.53 &0.69 &0.77 &0.90 &0.94 &0.97 &1.06  \\
Ours &\textbf{0.25}	&\textbf{0.43}	&\textbf{0.65}	&\textbf{0.75}	&\textbf{0.79}	&\textbf{0.85}	&\textbf{0.92}	&\textbf{0.96} \\
\hline
\hline
  

Methods  &Greeting & & & & & & & \\
\hline
    &80ms &160ms &320ms &400ms &560ms &640ms &720ms &1000ms\\
ERD \cite{fragkiadaki2015recurrent} &0.85 &1.09 &1.45 &1.64 &1.93 &1.89 &1.92 &1.98 \\
LSTM-3LR \cite{fragkiadaki2015recurrent}  &0.80 &0.99 &1.37 &1.54 &1.81 &1.76 &1.79 &1.85\\
Res-GRU \cite{martinez2017on}  &\textbf{0.52} &\textbf{0.86} &1.30 &1.47 &1.78 &1.75 &1.82 &1.96\\
Zero-velocity \cite{martinez2017on}  &0.54 &0.89 &1.30 &1.49 &1.79 &1.74 &1.77 &1.80 \\
MHU \cite{tang2018longterm}  &0.54 &0.87 &1.27 &\textbf{1.45} &1.75 &1.71 &1.74 &1.87 \\
Ours &\textbf{0.52}	&\textbf{0.86}	&\textbf{1.26}	&\textbf{1.45}	&\textbf{1.58}	&\textbf{1.69}	&\textbf{1.72}	&\textbf{1.79}  \\
\hline
\hline
Methods &Taking photo & & & & & & &  \\
\hline
    &80ms &160ms &320ms &400ms &560ms &640ms &720ms &1000ms\\
ERD \cite{fragkiadaki2015recurrent} &0.70 &0.78 &0.97 &1.09 &1.20 &1.23 &1.27 &1.37 \\
LSTM-3LR \cite{fragkiadaki2015recurrent} &0.63 &0.64 &0.86 &0.98 &1.09 &1.13 &1.17 &1.30\\
Res-GRU \cite{martinez2017on}  &0.29 &0.58 &0.90 &1.04 &1.17 &1.23 &1.29 &1.47\\
Zero-velocity \cite{martinez2017on}  &0.25 &0.51 &0.79 &0.92 &1.03 &\textbf{1.06} &\textbf{1.13} &1.27 \\
MHU \cite{tang2018longterm} &0.27 &0.54 &0.84 &0.96 &1.04 &1.08 &1.14 &1.35\\
Ours &\textbf{0.24}	&\textbf{0.50}	&\textbf{0.76}	&\textbf{0.89}	&\textbf{0.95}	&1.08	&1.15	&\textbf{1.24} \\
\hline
  \end{tabular}
  }
    \label{tab:main_3.6}
    \vspace{-1.cm}
\end{table*}

\subsection{Datasets}

\textbf{MultiTHUMOS:} The MultiTHUMOS dataset~\cite{yeung2015every} is a well-established video understanding dataset for multi-class activity prediction.
It contains 400 videos spanning about 30 hours of video and 65 action classes.

\textbf{Charades:} 
Charades~\cite{sigurdsson2016charades} is a challenging video dataset containing longer-duration activities recorded in home environments. Charades is a multi-class multi-label dataset in which multiple activities are often co-occurring. We use it to demonstrate the ability of the model to handle complex data.
It contains 9858 videos of 157 action classes.

\textbf{Human3.6M:} The Human 3.6M dataset~\cite{ionescu2014Human3.6} is a popular benchmark for future pose prediction. It has 3.6 million 3D human poses of 15 activities. 
The goal is to predict the future 3D locations of 32 joints in the human body. 

\textbf{50 Salads:} The 50 Salads \cite{50salads} is a video dataset of 50 salad preparation sequences (518,411 frames total) with an average length of 6.4 minutes per video. It has been used recently for future activity prediction \cite{ke2019future,farha2018when}, making it suitable for the evaluation of our method.

\subsection{Human Pose Forecasting}
We first evaluate the approach on forecasting 3D human pose, a real valued structured-output problem. This is a challenging task~\cite{jain2016structural,fragkiadaki2015recurrent} but is of high importance, e.g., for motion planning in robotics. It also showcases the use of the Adversarial Grammar, as using the standard grammar is not feasible due to the memory and computation constraints for this real-valued dataset.

\begin{table*} [t] 
\small
  \centering
    \caption{Evaluation of future pose for short-term and long-term prediction horizons. Measured with Mean Angle Error (lower is better) on Human3.6M. No predictions beyond 1 second are available for prior work.}
  {
  \begin{tabular}{l|c|c|c|c|c|c|c|c|c|c}
  \hline
  
\hline
Method    &80ms &160ms &320ms  &560ms &640ms &720ms &1s &2s &3s &4s\\
\hline
ERD \cite{fragkiadaki2015recurrent} &0.93 &1.07 &1.31  &1.58 &1.64 &1.70  &1.95 &- &- &-\\
LSTM-3LR \cite{fragkiadaki2015recurrent} &0.87 &0.93 &1.19  &1.49 &1.55 &1.62  &1.89 &- &- &-\\
Res-GRU \cite{martinez2017on} &0.40 &0.72 &1.09 &1.45 &1.52 &1.59  &1.89 &- &- &-\\
Zero-vel. \cite{martinez2017on} &0.40 &0.71 &1.07  &1.42 &1.50 &1.57  &1.85 &- &- &-\\
MHU-MSE \cite{tang2018longterm} &0.39 &0.69 &1.04  &1.40 &1.49 &1.57 &1.89 &- &- &-\\
MHU \cite{tang2018longterm} &0.39 &0.68 &1.01  &1.34 &1.42 &\textbf{1.49}  &1.80 &- &- &-\\
\hline
AGG (Ours) &\textbf{0.36}	&\textbf{0.65}	&\textbf{0.98}	&\textbf{1.27}	&\textbf{1.40}	&\textbf{1.49}	&\textbf{1.74}	&\textbf{2.25}	&\textbf{2.70}	&\textbf{2.98} \\


\hline
\hline

\hline
  \end{tabular}
  }
    \label{tab:main_3.6_mean}
\vspace{-0.5cm}
\end{table*}


\begin{figure}
  \centering
   \includegraphics[width=1.0\linewidth]{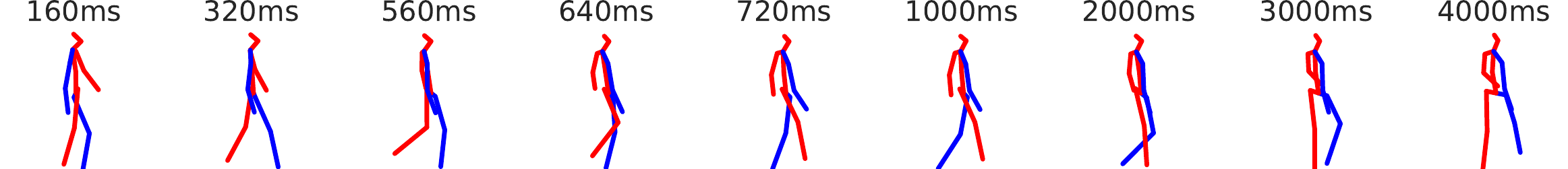}
   \includegraphics[width=1.0\linewidth]{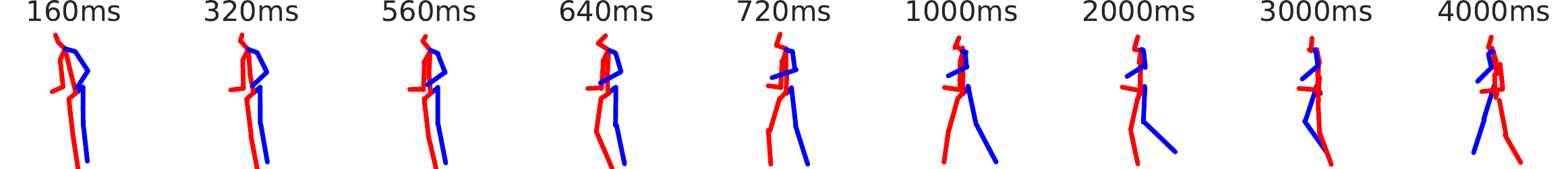}
   \includegraphics[width=1.0\linewidth]{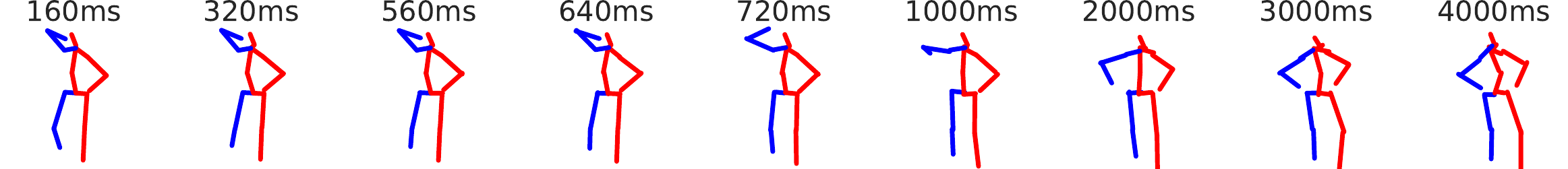}%
  \caption{Example results for 3D pose predictions. Top: walking, middle: greeting, bottom: posing.}
  \label{fig:3Dpose}
\end{figure}

\noindent\textbf{Human 3.6M dataset.} We conduct experiments on the well established future pose prediction benchmark Human3.6M ~\cite{ionescu2014Human3.6}. 
We here predict the future 3D locations of 32 joints in the human body. 
We use quaternions to represent each joint location, allowing for a more continuous joint representation space. We also predict differences, rather than absolute positions, which we found leads to more stable learning. 
Previous work demonstrated prediction results up to a second on this dataset. This work can generate future sequences for longer horizons, 4 seconds in the future. 

\begin{figure}[t]
  \centering
  \includegraphics[width=0.8\linewidth]{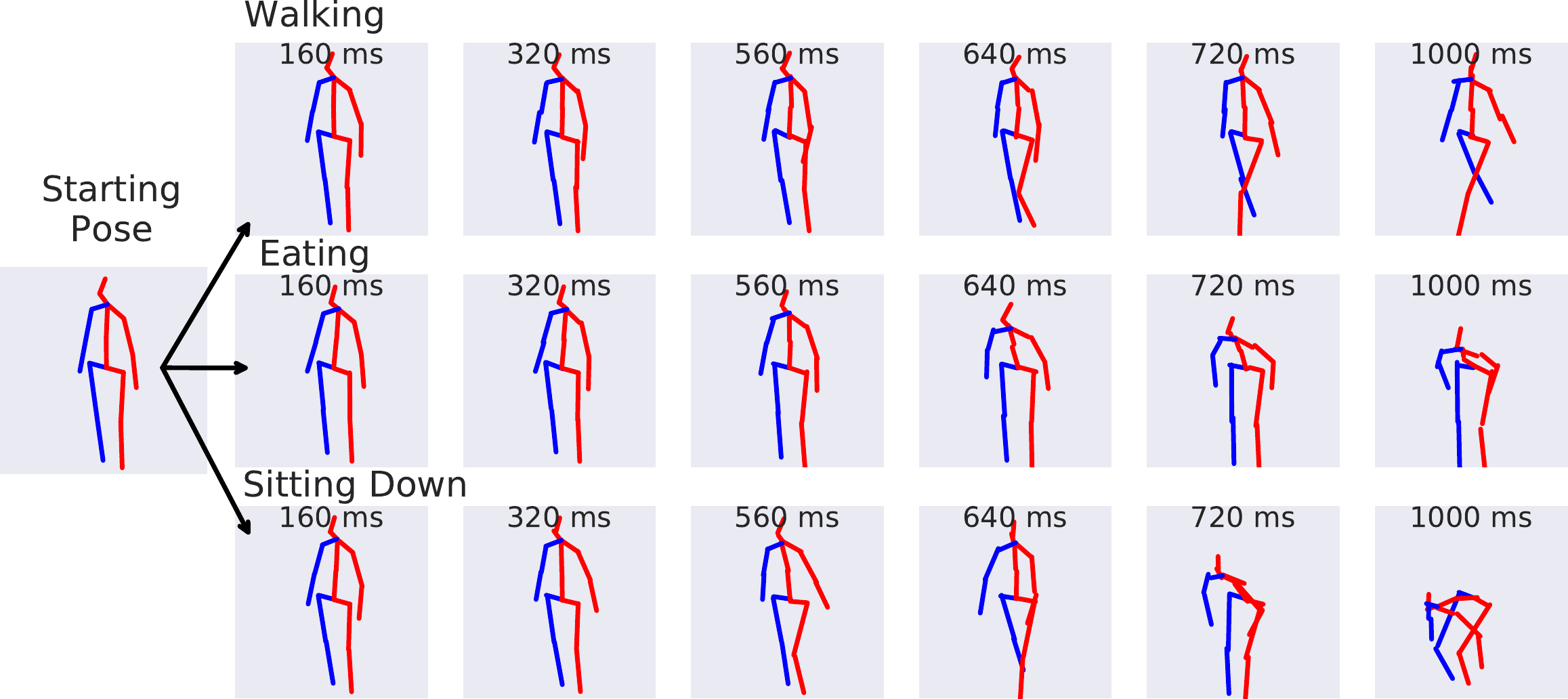}
  \caption{Starting from a neutral pose, the grammar is able to generate multiple sequences by selecting different rules. Top: a walking sequence, middle: eating, bottom: sitting.}
  \label{fig:pose-branch}
  \vspace{-0.2cm}
\end{figure}

We compare against the state-of-the-art methods on the Human 3.6M benchmark ~\cite{fragkiadaki2015recurrent,jain2016structural,ionescu2014Human3.6,martinez2017on,tang2018longterm} using the Mean Angle Error (MAE) metric as introduced by \cite{jain2016structural}.
Table~\ref{tab:main_3.6} shows results on several activities and
Table~\ref{tab:main_3.6_mean} shows average MAE for all activities compared to the state-of-the-art methods, 
consistent with the protocol in prior work.
As seen from the tables, our work outperforms prior work. Furthermore, we are able to generate results at larger time horizons of four seconds in the future.
In Figure~\ref{fig:3Dpose}, we show some predicted future poses for several different activities, confirming the results reflect the characteristics of the actual behaviors. In Figure~\ref{fig:pose-branch}, we show the ability of the adversarial grammar to generate different sequences from a given starting state. Here, given the same starting state, we select different rules, which lead to different sequences corresponding to walking, eating or sitting.









\subsection{Activity forecasting in videos}
We further test the method for video activity anticipation, where the goal is to predict future activities at various time-horizons, using an initial video sequence as input. We predict future activities on three video understanding datasets MultiTHUMOS~\cite{yeung2015every},  Charades~\cite{sigurdsson2016charades} and 50-salads \cite{50salads} using the standard evaluation protocols per dataset.
 We also predict from 1 to 45 seconds in the future on MultiTHUMOS and Charades, which is much further into the future than prior approaches.


\vspace{-5pt}
\subsubsection{50 Salads.}
\begin{figure}[t]
    \centering
    \includegraphics[width=0.8\linewidth]{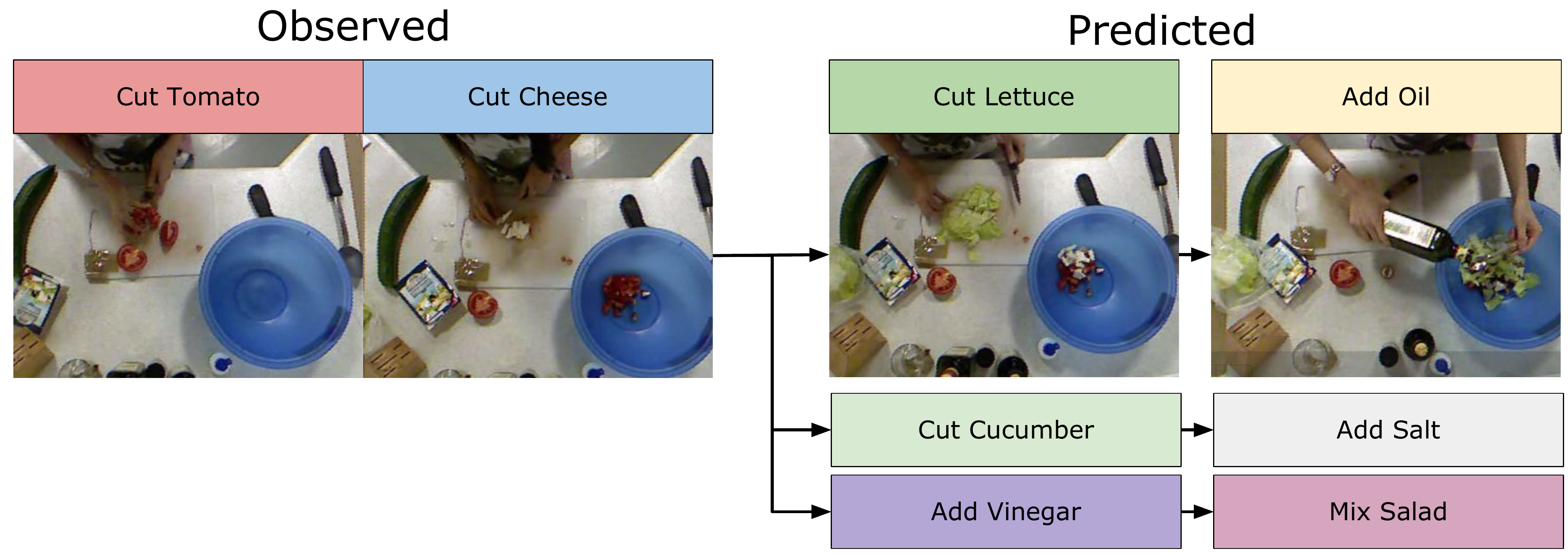}
    \caption{Example sequence from 50-salads showing the observed frames and the next two predictions.}
    \label{fig:salad-figure}
\end{figure}

Following the setting `without ground truth' in \cite{ke2019future} and~\cite{farha2018when}, we evaluate the future prediction task on the 50 Salads dataset \cite{50salads}. As per standard evaluation protocol, we report prediction on portions of the video when $20\%$ and $30\%$ portion is observed.
The results are shown in Table \ref{tab:salad}, where Grammar-only denotes training without adversarial losses. The results confirm that our approach allows better prediction which outperforms both the baseline, which is already a strong grammar model, as well as, the state-of-the-art approaches. Fig.~\ref{fig:salad-figure} has an example prediction, which proposes three plausible continuations of the recipe, the top corresponding to the ground truth.

\begin{table}[]
    \centering
    \vspace{-8mm}
    \caption{Results on 50 Salads without ground-truth observations. The proposed work outperforms the grammar baselines and the state-of-the-art. $^*$ measured using mean-over-classes. $^\dagger$ measured using average per-frame accuracy. 
    }
    \begin{tabular}{l@{\hspace{4pt}}c@{\hspace{4pt}}c@{\hspace{4pt}}c@{\hspace{4pt}}c@{\hspace{4pt}}c@{\hspace{4pt}}c@{\hspace{4pt}}c@{\hspace{4pt}}c@{\hspace{4pt}}}
    \toprule
      Observation & \multicolumn{4}{c}{20\%} & \multicolumn{4}{c}{30\%} \\
      \cmidrule(lr){2-5}\cmidrule(lr){6-9}
      Prediction & 10\% & 20\% & 30\% & 50\% & 10\% & 20\% & 30\% & 50\% \\
      \midrule
      Nearest-Neighbor~\cite{farha2018when}$^*$ &19.0 &16.1 &14.1 &10.4 &21.6 &15.5 &13.5 &13.9 \\
      RNN~\cite{farha2018when}$^*$               & 30.1 & 25.4 & 18.7 & 13.5 & 30.8 & 17.2 & 14.8 & 9.8 \\
      CNN~\cite{farha2018when} $^*$              & 21.2 & 19.0 & 16.0 & 9.9 & 29.1 & 20.1 & 17.5 & 10.9 \\
TCA \cite{ke2019future}$^*$ & 32.5 & 27.6 & 21.3 & 16.0 & 35.1 & 27.1 & 22.1 & 15.6\\
 Grammar (from~\cite{farha2018when})$^*$ &24.7 &22.3 &19.8 &12.7 &29.7 &19.2 &15.2 &13.1 \\     
\hline
    Grammar only$^*$      & 40.2 & 33.1 & 24.1 & 18.7 & 39.2 & 30.1 & 25.2 & 17.9 \\
    AGG (Ours)$^*$       & \textbf{40.4} & \textbf{33.7} & \textbf{25.4} & \textbf{20.9} & \textbf{40.7} & \textbf{40.1} & \textbf{26.4} & \textbf{19.2} \\
    \midrule
    Grammar only$^\dagger$      & 39.2 & 32.1 & 24.8 & 19.3 & 38.4 & 29.5 & 25.5 & 18.5 \\
    AGG (Ours)$^\dagger$       & {39.5} & {33.2} & {25.9} & {21.2} & {39.5} & {31.5} & {26.4} & {19.8} \\
      \bottomrule
    \end{tabular}
    \label{tab:salad}
    \vspace{-0.5cm}
\end{table}


\vspace{-5pt}
\subsubsection{MultiTHUMOS.}
We here present our future prediction results on the MultiTHUMOS dataset~\cite{yeung2015every} \footnote{Note that most of the previous works used the MultiTHUMOS dataset and the Charades dataset for per-frame activity categorization; our works showcases a long-term activity forecasting capability, instead. 
}.
We use a standard evaluation metric: we predict the activities occurring $T$ seconds in the future and compute the mean average precision (mAP) between the predictions and ground truth. As the grammar model is able to generate multiple, different future sequences, we also report the maximum mAP the model could obtain by selecting the best of 10 different future predictions. We compare the predictions at 1, 2, 5, 10, 20, 30 and 45 seconds into the future.
As little work has explored long-term future activity prediction (with the exception of ~\cite{yeung2015every} which predicts within a second), we compare against four different baseline methods: (i) repeating the activity prediction of the last seen frame, (ii) using a fully connected layer to predict the next second (applied autoregressively), (iii) using a fully-connected layer to directly predict activities at various future times, and (iv) an LSTM applied autoregressively to future activity predictions.

Table~\ref{tab:thumos} shows activity prediction accuracy for the MultiTHUMOS dataset. 
In the table, we also report our approach when limited to generating a single outcome (`AGG-single'), to be consistent to previous methods which are not able to generate more than one outcome. We also compare to grammar without adversarial learning, trained by pruning the exponential amount of future sequences to fit into the memory (`Grammar only`).

\begin{table*} 
  \centering
 \caption{Prediction mAP for future activities (higher is better) from 1 seconds to 45 seconds in the future. MultiTHUMOS.}
   
  {
   
  \begin{tabular}{l|c|c|c|c|c|c|c}
  \hline
  
\hline
Method    &1 sec	&2 sec	&5 sec	&10 sec	&20 sec	&30 sec	&45 sec\\
\hline

Random	&2.6	&2.6	&2.6	&2.6	&2.6	&2.6	&2.6 \\
Last Predicted Action	&16.5	&16.0	&15.1	&12.7	&8.7	&5.8	&5.9 \\
FC Autoregressive	&17.9	&17  	&14.5	&7.7	&4.5	&4.2	&4.7\\
FC Direct	&13.7	&9.8	&11.0	&7.3	&8.0	&5.5	&8.2\\
LSTM (Autoregressive)	&16.5	&15.7	&12.5	&6.8	&4.1	&3.2	&3.9\\
\hline
Grammar only 	&18.7	&18.6	&13.5	&12.8	&10.5	&8.2	&8.5\\
AGG-single (Ours)	&19.3	&19.6	&13.1	&13.6	&11.7	&10.4	&\textbf{11.4}\\
AGG (Ours)	&\textbf{22.0}	&\textbf{19.9}	&\textbf{15.5}	&\textbf{14.4}	&\textbf{13.3}	&\textbf{10.8}	&\textbf{11.4}\\
\hline
  \end{tabular}
  }
    \label{tab:thumos}
\end{table*}

\begin{figure}[t] 
    \centering
    \includegraphics[width=0.6\linewidth]{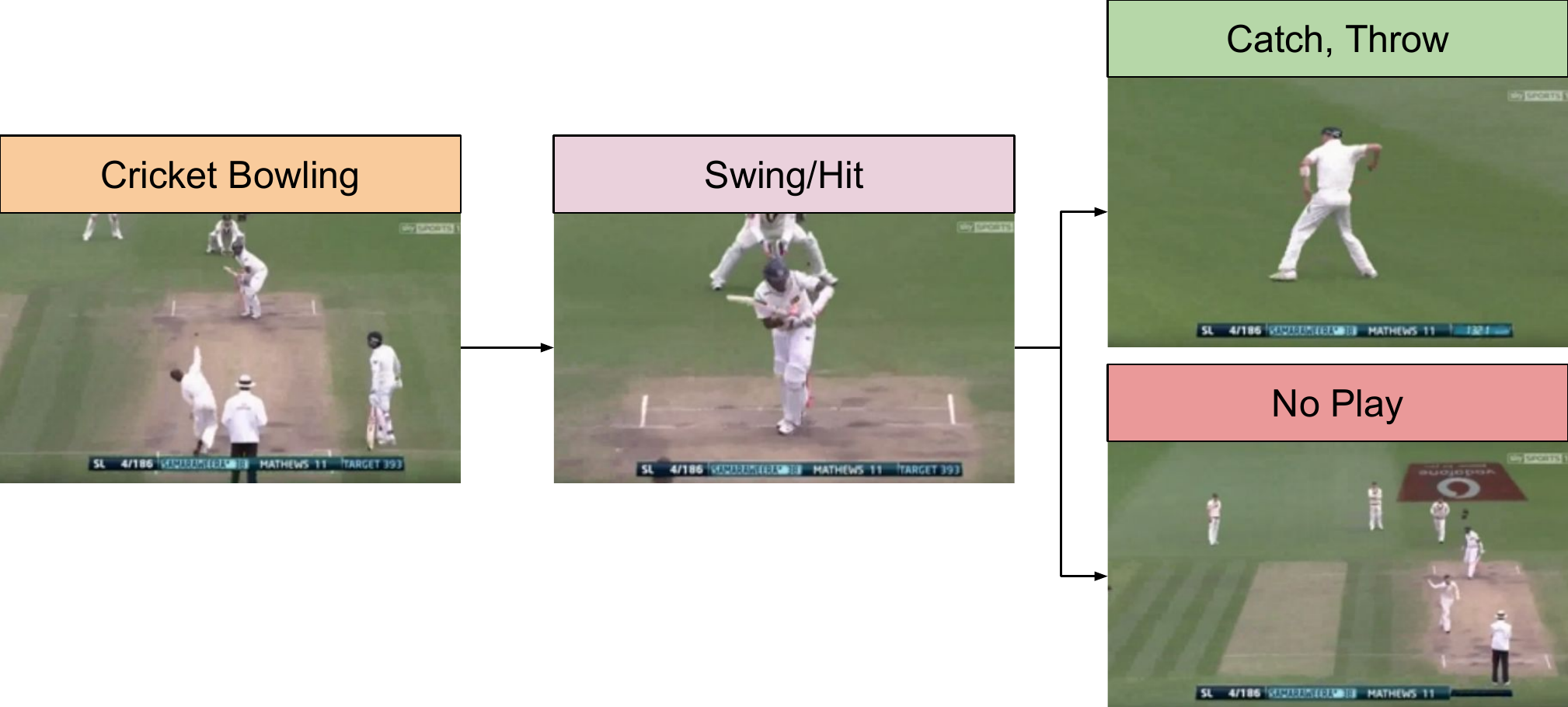}
    \caption{Example video and activity sequence from MultiTHUMOS (a cricket game). The adversarial grammar is able to learn two possible sequences: a hit/play and no play, instead of picking only the most likely one.}
    \label{fig:cricket}
    \vspace{-4mm}
\end{figure}

As seen, our approach outperforms alternative methods. We observe that the gap to other approaches widens further in the future: 3.9 mAP for the LSTM vs 11.2 of ours at 45 sec. in the future, as the autoregressive predictions of an LSTM become noisy. Due to the structure of the grammar model, we are able to generate better long-term predictions. We also find that by predicting multiple futures and taking the max improves performance, confirming that the grammar model is generating different sequences, some of which more closely match the ground truth (see also Figure ~\ref{fig:cricket}).



\vspace{-5pt}
\subsubsection{Charades.}

Table~\ref{tab:charades} shows the future activity prediction results on Charades, using the same protocol as MultiTHUMOS. 
Similar to our MultiTHUMOS experiments, we observe that the adversarial grammar model provides more accurate future prediction than previous work, outperforming the grammar-only model in most cases. While the grammar-only model performs slightly better at 10 and 20 seconds, it is not computationally feasible for real-valued tasks due to the memory constraint.
We note that Charades is more challenging than others on both recognition and prediction. 
Figure~\ref{fig:activity} shows a true sequence and several other sequences generated by the adversarial grammar. As Charades contains many different possible sequences, generating multiple  futures is beneficial.

\begin{table*} 
  \centering
  \caption{Prediction accuracy for future activities for 45 seconds in the future on the Charades dataset. }
    
  {

  \begin{tabular}{l|c|c|c|c|c|c|c}
  \hline
  
\hline
Method    &1 sec	&2 sec	&5 sec	&10 sec	&20 sec	&30 sec	&45 sec\\
\hline

Random	&2.4	&2.4	&2.4	&2.4	&2.4	&2.4	&2.4 \\
Last Predicted Action	&15.1	&13.8	&12.8	&10.2	&7.6	&6.2	&5.7\\
FC Autoregressive	&13.5	&14.0	&12.6	&6.7	&3.7	&3.5	&5.1\\
FC Direct	&15.2	&14.5	&12.2	&9.1	&6.6	&6.5	&5.5\\
LSTM (Autoregressive) 	&12.6	&12.7	&12.4	&10.8	&7.0	&6.1	&5.4\\
\hline
Grammar only &15.7	&14.8	&12.9	&\textbf{11.2}	&\textbf{8.5}	&6.6	&8.5\\
AGG-single  (Ours)	&15.9	&15.0	&13.1	&10.5	&7.4	  &6.2	&8.8\\
AGG  (Ours) 	 &\textbf{17.0}	&\textbf{15.9}	&\textbf{13.4}	&10.7	&7.8	&\textbf{7.2}	&\textbf{9.8}\\
\hline
  \end{tabular}
  }
    \label{tab:charades}
\end{table*}





\vspace{-5pt}
\subsubsection{Ablation study}

We conduct additional experiments to examine the importance of learning grammar  with multiple possibilities (i.e., branching). Table \ref{tab:ablation} compares the models with and without the branching capability. These models use the exact same network architecture as our full models, while the only difference is that they do not generate multiple possible sequences for its learning. That is, they just become standard RNNs, constrained to have our grammar structure. We are able to observe that the ability to consider multiple possibilities during the learning is important, and that our adversarial training is beneficial. Note that we restricted these models to only generate one sequence with the highest likelihood during the inference for fair comparison.

\vspace{-0.5cm}
\begin{table} [h!]
\small
  \centering
    \caption{Ablation of our grammar learning on Charades.}
  {

  \begin{tabular}{l|c|c|c}
  \hline
  
\hline
Method    &1 sec	&5 sec	&45 sec\\
\hline
Grammar only - no branching   & 12.2   &8.4  &3.8\\
Grammar only 	&15.7 &12.9	&8.5\\
\hline
Adversarial Grammar (AGG) - no branching    & 14.2  &12.5 &5.5  \\
Adversarial Grammar (AGG) 	&\textbf{15.9} &\textbf{13.1}	&\textbf{8.8}\\
\hline
  \end{tabular}
  }
    \label{tab:ablation}
\end{table}

\section{Conclusion}
We proposed a differentiable adversarial generative grammar which shows strong performance for future prediction of human pose and activities.
Because of the structure we impose for learning grammar-like rules for sequences and learning in adversarial fashion, the model is able to generate multiple sequences that follow the distribution seen in data. One challenge is evaluating future predictions when the ground truth only contains one of many potentially valid sequences. In the future, other forms of evaluation, such as asking humans to rate a generated sequence, could be explored.



{\small
\bibliographystyle{splncs04}
\bibliography{egbib}

\begin{thebibliography}{10}
\providecommand{\url}[1]{\texttt{#1}}
\providecommand{\urlprefix}{URL }
\providecommand{\doi}[1]{https://doi.org/#1}

\bibitem{babaeizadeh2017stochastic}
Babaeizadeh, M., Finn, C., Erhan, D., Campbell, R.H., Levine, S.: Stochastic
  variational video prediction. arXiv preprint arXiv:1710.11252  (2017)

\bibitem{brock2019large}
Brock, A., Donahue, J., Simonyan, K.: Large scale gan training for high
  fidelity natural image synthesis. ICLR  (2019)

\bibitem{carreira2017quo}
Carreira, J., Zisserman, A.: Quo vadis, action recognition? a new model and the
  kinetics dataset. In: CVPR (2017)

\bibitem{cho2014GRU}
Cho, K., van Merrienboer, B., Gulcehre, C., Bahdanau, D., Bougares, F.,
  Schwenk, H., Bengio, Y.: Learning phrase representations using rnn
  encoder-decoder for statistical machine translation. EMNLP  (2014)

\bibitem{chomsky1956three}
Chomsky, N.: Three models for the description of language. IRE Transactions on
  information theory  \textbf{2}(3),  113--124 (1956)

\bibitem{denton2018stochastic}
Denton, E., Fergus, R.: Stochastic video generation with a learned prior. arXiv
  preprint arXiv:1802.07687  (2018)

\bibitem{denton2015deep}
Emily L~Denton, Soumith~Chintala, R.F.: Deep generative image models using a
  laplacian pyramid of adversarial networks. Advances in Neural Information
  Processing Systems (NeurIPS)  (2015)

\bibitem{farha2018when}
Farha, Y.A., Richard, A., Gall, J.: When will you do what? - anticipating
  temporal occurrences of activities. In: CVPR (2018)

\bibitem{fedus2018maskgan}
Fedus, W., Goodfellow, I., Dai, A.: Maskgan: Better text generation via filling
  in the $\_$. ICLR  (2018)

\bibitem{finn2016unsupervised}
Finn, C., Goodfellow, I., Levine, S.: Unsupervised learning for physical
  interaction through video prediction. In: Advances in Neural Information
  Processing Systems (NeurIPS). pp. 64--72 (2016)

\bibitem{fraccaro2016sequential}
Fraccaro, M., S{\o}nderby, S.K., Paquet, U., Winther, O.: Sequential neural
  models with stochastic layers. In: Advances in neural information processing
  systems. pp. 2199--2207 (2016)

\bibitem{fragkiadaki2015recurrent}
Fragkiadaki, K., Levine, S., Felsen, P., Malik, J.: Recurrent network models
  for human dynamics. In: ICCV (2015)

\bibitem{goodfellow2014gans}
Goodfellow, I., Pouget-Abadie, J., Mirza, M., Xu, B., WardeFarley, D., Ozair,
  S., Courville, A., Bengio, Y.: Generative adversarial nets. Advances in
  Neural Information Processing Systems (NeurIPS)  (2014)

\bibitem{han2008bottom}
Han, F., Zhu, S.C.: Bottom-up/top-down image parsing with attribute grammar.
  IEEE Transactions on Pattern Analysis and Machine Intelligence
  \textbf{31}(1),  59--73 (2008)

\bibitem{hu2017toward}
Hu, Z., Yang, Z., Liang, X., Salakhutdinov, R., Xing, E.P.: Toward controlled
  generation of text. ICML  (2017)

\bibitem{ionescu2014Human3.6}
Ionescu, C., Papava, D., Olaru, V., Sminchisescu, C.: Human3.6m: Large scale
  datasets and predictive methods for 3d human sensing in natural environments.
  IEEE Transactions on Pattern Analysis and Machine Intelligence  (2014)

\bibitem{isola2017i2i}
Isola, P., Zhu, J.Y., Zhou, T., Efros, A.A.: Image-to-image translation with
  conditional adversarial networks. CVPR  (2017)

\bibitem{jain2016structural}
Jain, A., Zamir, A.R., Savarese, S., Saxena, A.: Structural-rnn: Deep learning
  on spatio-temporal graphs. In: CVPR (2016)

\bibitem{jang2016categorical}
Jang, E., Gu, S., Poole, B.: Categorical reparameterization with
  gumbel-softmax. In: ICLR (2017)

\bibitem{ke2019future}
Ke, Q., Fritz, M., Schiele, B.: Time-conditioned action anticipation in one
  shot. In: CVPR (2019)

\bibitem{lee2018stochastic}
Lee, A.X., Zhang, R., Ebert, F., Abbeel, P., Finn, C., Levine, S.: Stochastic
  adversarial video prediction. arXiv preprint arXiv:1804.01523  (2018)

\bibitem{maddison2016concrete}
Maddison, C.J., Mnih, A., Teh, Y.W.: The concrete distribution: A continuous
  relaxation of discrete random variables. In: ICLR (2017)

\bibitem{martinez2017on}
Martinez, J., Black, M., , Romero, J.: On human motion prediction using
  recurrent neural networks. In: CVPR (2017)

\bibitem{moore2002recognizing}
Moore, D., Essa, I.: Recognizing multitasked activities from video using
  stochastic context-free grammar. In: Proceedings of AAAI Conference on
  Artificial Intelligence (AAAI). pp. 770--776 (2002)

\bibitem{pirsiavash2014parsing}
Pirsiavash, H., Ramanan, D.: Parsing videos of actions with segmental grammars.
  In: CVPR. pp. 612--619 (2014)

\bibitem{qi2017predicting}
Qi, S., Huang, S., Wei, P., Zhu, S.C.: Predicting human activities using
  stochastic grammar. In: Proceedings of the IEEE International Conference on
  Computer Vision. pp. 1164--1172 (2017)

\bibitem{qi2018parser}
Qi, S., Jia, B., Zhu, S.C.: Generalized earley parser: Bridging symbolic
  grammars and sequence data for future prediction. arXiv preprint
  arXiv:1806.03497  (2018)

\bibitem{ryoo2006recognition}
Ryoo, M.S., Aggarwal, J.K.: Recognition of composite human activities through
  context-free grammar based representation. In: 2006 IEEE Computer Society
  Conference on Computer Vision and Pattern Recognition (CVPR'06). vol.~2, pp.
  1709--1718. IEEE (2006)

\bibitem{sigurdsson2016charades}
Sigurdsson, G.A., Varol, G., Wang, X., Farhadi, A., Laptev, I., Gupta, A.:
  Hollywood in homes: Crowdsourcing data collection for activity understanding.
  Proceedings of European Conference on Computer Vision (ECCV)  (2016)

\bibitem{socher2011parsing}
Socher, R., Lin, C.C., Manning, C., Ng, A.Y.: Parsing natural scenes and
  natural language with recursive neural networks. In: Proceedings of the 28th
  international conference on machine learning (ICML-11). pp. 129--136 (2011)

\bibitem{50salads}
Stein, S., McKenna, S.J.: Combining embedded accelerometers with computer
  vision for recognizing food preparation activities. In: Proceedings of the
  2013 ACM international joint conference on Pervasive and ubiquitous
  computing. pp. 729--738. ACM (2013)

\bibitem{tang2018longterm}
Tang, Y., Ma, L., Liu, W., Zheng, W.S.: Long-term human motion prediction by
  modeling motion context and enhancing motion dynamic. In: IJCAI (2018)

\bibitem{vo2014stochastic}
Vo, N.N., Bobick, A.F.: From stochastic grammar to bayes network: Probabilistic
  parsing of complex activity. In: Proceedings of the IEEE Conference on
  Computer Vision and Pattern Recognition. pp. 2641--2648 (2014)

\bibitem{wang2018pix2pixHD}
Wang, T.C., Liu, M.Y., Zhu, J.Y., Tao, A., Kautz, J., Catanzaro, B.:
  High-resolution image synthesis and semantic manipulation with conditional
  gans. In: CVPR (2018)

\bibitem{yang2017differentiable}
Yang, F., Yang, Z., Cohen, W.W.: Differentiable learning of logical rules for
  knowledge base reasoning. Advances in Neural Information Processing Systems
  (NeurIPS)  (2017)

\bibitem{yeung2015every}
Yeung, S., Russakovsky, O., Jin, N., Andriluka, M., Mori, G., Fei-Fei, L.:
  Every moment counts: Dense detailed labeling of actions in complex videos.
  International Journal of Computer Vision (IJCV) pp. 1--15 (2015)

\bibitem{yagatama2018memory}
Yogatama, D., Miao, Y., Melis, G., Ling, W., Kuncoro, A., Dyer, C., Blunsom,
  P.: Memory architectures in recurrent neural network language models. ICLR
  (2018)

\bibitem{yu2017seqgan}
Yu, L., Zhang, W., J.Wang, Yu, Y.: Seqgan: sequence generative adversarial nets
  with policy gradient. Proceedings of AAAI Conference on Artificial
  Intelligence (AAAI)  (2017)

\bibitem{zhao2011image}
Zhao, Y., Zhu, S.C.: Image parsing with stochastic scene grammar. In: Advances
  in Neural Information Processing Systems. pp. 73--81 (2011)

\bibitem{zhu2007stochastic}
Zhu, S.C., Mumford, D.: A stochastic grammar of images. Foundations and
  Trends® in Computer Graphics and Vision  \textbf{2} (2007)

\end{thebibliography}
}

\clearpage
\newpage
\appendix

\section{Implementation Details}

\paragraph{Activity Prediction} For activity prediction, the number of non-terminals ($\mathcal{N}$) was set to 64, the number of terminals ($\mathcal{T}$) was set to the number of classes in the dataset (e.g., 65 in MultiTHUMOS and 157 in Charades). We used 4 rules for each non-terminal (a total of 256 rules). $G$, $f_R$, $f_T$ and $f_N$ each used one fully connected layer with sizes matching the desired inputs/outputs. $s$ is implemented as a two sequential temporal convolutional layers with 512 channels, followed by mean-pooling and a fully-connected layer to generate $N_0$.

\paragraph{3D Pose prediction} For 3D pose prediction, the number of non-terminals ($\mathcal{N}$) was set to 1024, the number of terminals ($\mathcal{T}$) was set to 1024, where each terminal has size of 128 (32 joints in 4D quaternion representation). The number of rules was set to 2 per non-terminal (a total of 2048 rules). $G$ was composed of 2 fully connected layers, $f_R$, $f_T$ and $f_N$ each used three fully connected layers with sizes matching the desired inputs/outputs. $s$ was implemented as a 2-layer GRU using a representation size of 1024, followed by mean-pooling and a fully-connected layer to generate $N_0$.

\subsection{Network Architecture}
Here we provide full details on the structure of the networks.

\paragraph{CNN for Starting Non-terminal} The function $s$ (from Eq. 4) is implemented using I3D \cite{carreira2017quo}. The input to the network is multiple frames with size $224\times 224$. The number of frames varies based on how many seconds of video is shown to the network before future prediction. This is at least 16 frames and at most 256 frames. This feature is then used as input to the temporal convolution or GRU described above.

\paragraph{Discriminator Architecture} The structure of $D$ is relatively simple. We use 3 1D convolutional layers with a kernel size of 5 and a stride of 4. This gives a temporal receptive field size of 84, which captures long temporal durations (at 12fps, this is 7 seconds per-feature). These layers have 128, 256, and then 64 channels. This is followed by mean-pooling to obtain the feature used for binary classification by a fully-connected layer.

We also tried using an RNN for the discriminator, but found it had comparable performance, but was slower during training.

\paragraph{Training Details} The model is trained for 5000 iterations using gradient descent with momentum of 0.9 and the initial learning rate set to 0.1. We follow the cosine learning rate decay schedule. Our implementation is in PyTorch and our models were trained on a single V100 GPU.

\section{Supplemental results}

\begin{figure*}
  \centering
   \includegraphics[width=\linewidth]{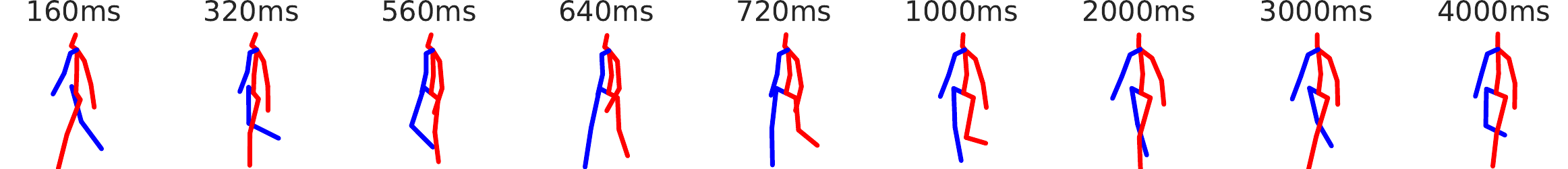}
   \includegraphics[width=\linewidth]{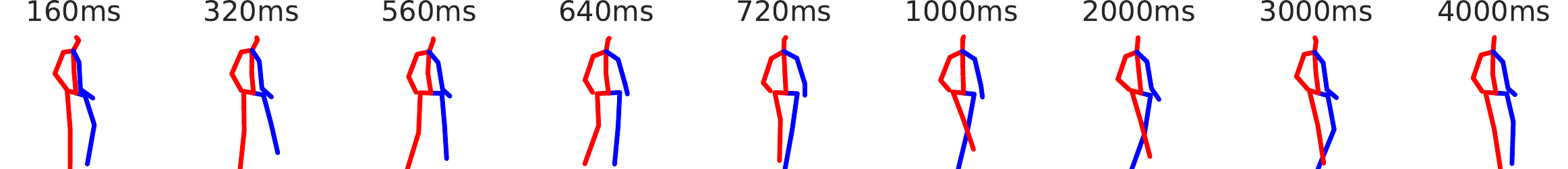}
   \includegraphics[width=\linewidth]{images/pose_greeting_1.pdf}
   \includegraphics[width=\linewidth]{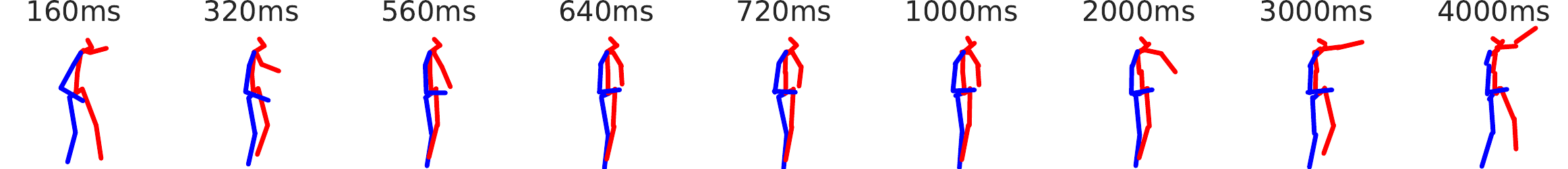}
   \includegraphics[width=\linewidth]{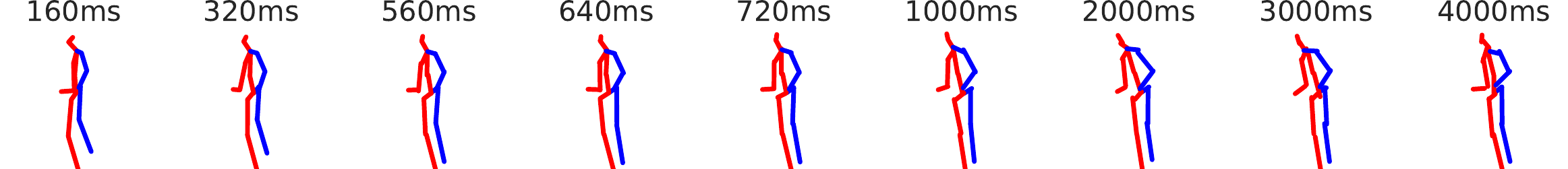}
   \includegraphics[width=\linewidth]{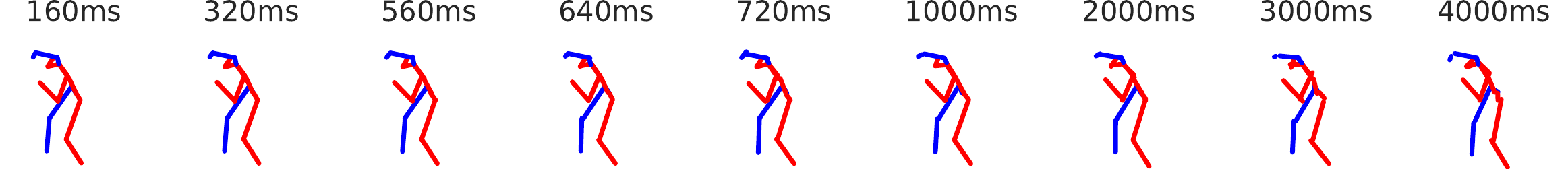}
   \includegraphics[width=\linewidth]{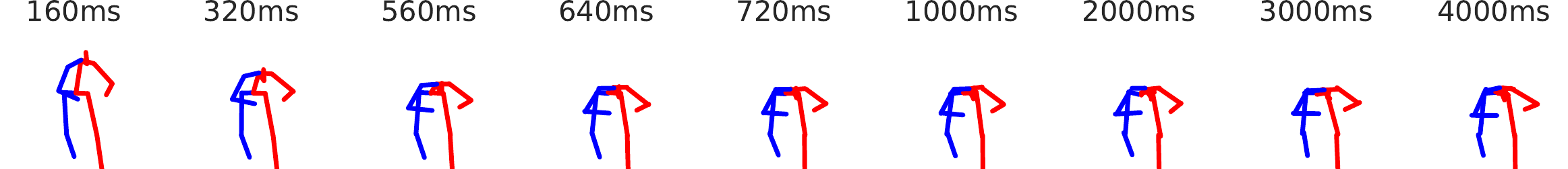}
   \includegraphics[width=\linewidth]{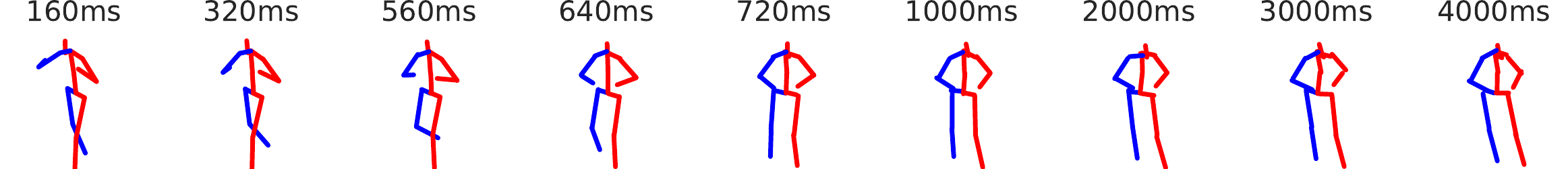}
   \includegraphics[width=\linewidth]{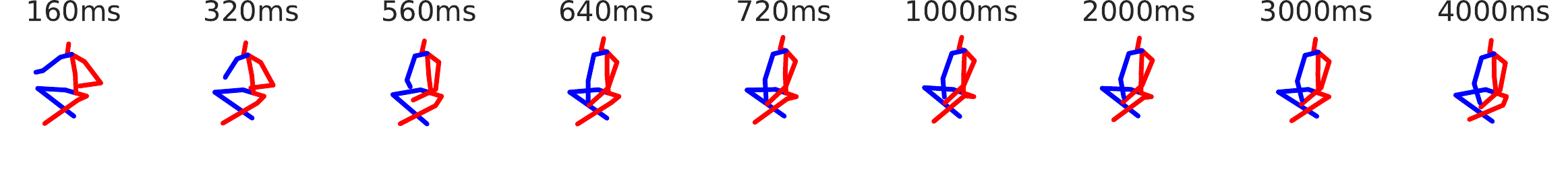}
   \includegraphics[width=\linewidth]{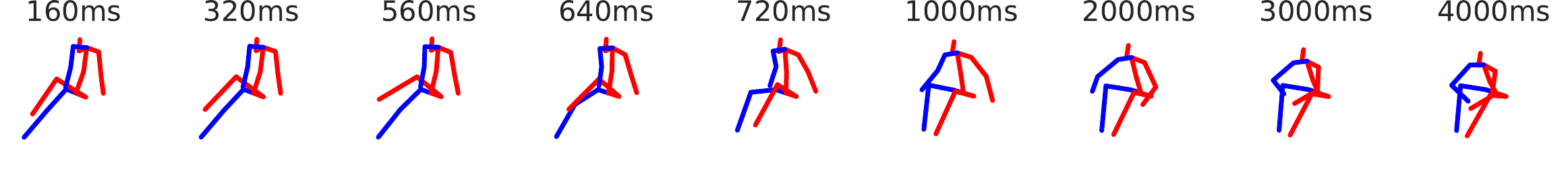}
   \includegraphics[width=\linewidth]{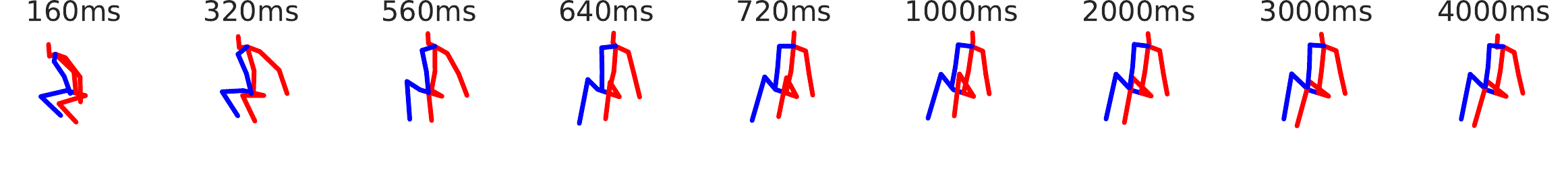}
   \includegraphics[width=\linewidth]{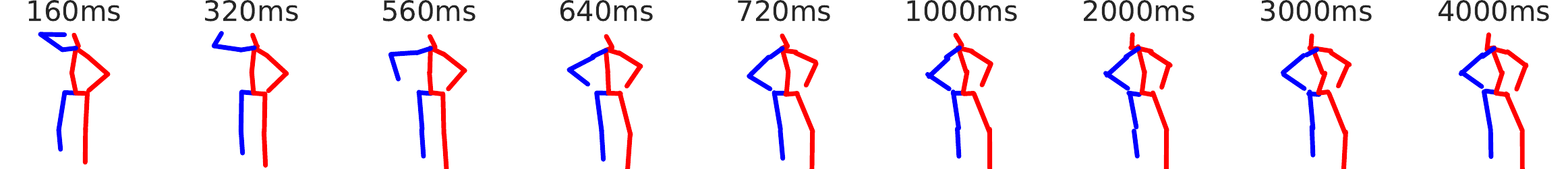}
   \includegraphics[width=\linewidth]{images/pose_posing_2.pdf}
  \caption{Various predicted 3D pose sequences for walking, greeting, taking photos, sitting, posing.}
  \label{fig:morepose}
\end{figure*}

Table~\ref{tab:main_3.6_mean} provides results of our approach for future 3D human pose prediction for all activities in the Human3.6M dataset. 
Figure~\ref{fig:morepose} shows more examples of future predicted 3D pose at different timesteps. 

\begin{table*} [h!]
\small
  \centering
  {
  \begin{tabular}{|l|c|c|c|c|c|c|c|c|c|c|c|}
  \hline
  
\hline

Activity	&80ms	&160ms	&320ms	&400ms	&560ms	&640ms	&720ms	&1s &2s &3s &4s \\
\hline
Walking	&0.25	&0.43	&0.65	&0.75	&0.79	&0.85	&0.92	&0.96	&1.37	&1.34	&1.87 \\
Eating	&0.2	&0.34	&0.53	&0.67	&0.79	&0.92	&1.01	&1.23	&1.66	&2.01	&2.14\\
Smoking	&0.26	&0.49	&0.92	&0.89	&0.99	&1.01	&1.02	&1.25	&1.95	&2.8	&3.37\\
Discussion	&0.29	&0.65	&0.91	&1.00	&1.23	&1.52	&1.68	&1.93	&2.32	&2.58	&2.65\\
Directions	&0.39	&0.59	&0.78	&0.87	&0.99	&1.01	&1.25	&1.46	&1.88	&2.37	&2.19\\
Greeting	&0.52	&0.86	&1.26	&1.45	&1.58	&1.69	&1.72	&1.79	&2.56	&3.08	&2.3\\
Phoning	&0.59	&1.15	&1.51	&1.65	&1.47	&1.71	&1.78	&1.84	&2.63	&2.97	&3.71\\
Posing	&0.25	&0.54	&1.19	&1.43	&1.86	&2.10	&2.15	&2.66	&3.46	&4.04	&4.49\\
Purchases	&0.6	&0.85	&1.16	&1.23	&1.58	&1.67	&1.72	&2.4	&1.95	&2.35	&2.63\\
Sitting	&0.39	&0.62	&1.02	&1.17	&1.24	&1.42	&1.48	&1.65	&2.73	&3.09	&3.47\\
SittingDown	&0.39	&0.75	&1.10	&1.23	&1.35	&1.48	&1.65	&1.88	&2.71	&3.88	&4.81\\
TakePhoto	&0.24	&0.5	&0.76	&0.89	&0.95	&1.08	&1.15	&1.24	&2.1	&2.45	&2.72\\
Waiting	&0.31	&0.61	&1.13	&1.37	&1.75	&1.92	&2.12	&2.55	&2.82	&3.18	&3.53\\
WalkingDog	&0.54	&0.87	&1.19	&1.35	&1.62	&1.75	&1.82	&1.91	&2.18	&2.83	&2.77\\
WalkTogether	&0.25	&0.51	&0.7	&0.74	&0.82	&0.88	&0.91	&1.33	&1.4	&1.62	&2.14\\
\hline									
Average &0.36	&0.65	&0.98	&1.11 &1.27	&1.40	&1.49	&1.74	&2.25	&2.70	&2.98 \\

\hline
  \end{tabular}
  }
  \caption{Evaluation of future pose of our appoach for both short-term and long-term prediction horizons for all activities. Human3.6M benchmark.}
    \label{tab:main_3.6_mean2}
\end{table*}

\end{document}